\pdfoutput=1

\documentclass[11pt]{article}

\usepackage[]{acl}

\usepackage{times}
\usepackage{latexsym}
\usepackage[T1]{fontenc}
\usepackage[utf8]{inputenc}
\usepackage{microtype}
\usepackage{inconsolata}

\usepackage{graphicx}
\usepackage{amssymb}
\usepackage{booktabs} 
\usepackage{multirow}
\usepackage{mathrsfs}
\usepackage{enumitem}
\usepackage{caption}
\usepackage{subcaption}
\usepackage{bm}
\usepackage{diagbox}
\usepackage{bbm}
\usepackage{amsmath}
\usepackage{algorithm}
\usepackage{algorithmicx}
\usepackage{algpseudocodex}
\usepackage[capitalize]{cleveref}
\usepackage{array}
\usepackage{colortbl}
\usepackage{amsfonts} 

\usepackage[utf8]{inputenc} 
\usepackage[T1]{fontenc}    
\usepackage{hyperref}       
\usepackage{url}            
\usepackage{booktabs}       
\usepackage{amsfonts}       
\usepackage{nicefrac}       

\usepackage{microtype}      
\usepackage{xcolor}  

\usepackage{inconsolata}

\usepackage[utf8]{inputenc}
\usepackage[T1]{fontenc} 

\crefname{section}{Sec.}{Secs.}
\Crefname{section}{Section}{Sections}

\usepackage{colortbl}
\usepackage{hyperref}
\definecolor{mygray}{gray}{.9}
\definecolor{lightgray}{gray}{0.92}

\usepackage[T1]{fontenc}

\usepackage[utf8]{inputenc}

\usepackage{microtype}

\crefname{section}{Sec.}{Secs.}
\Crefname{section}{Section}{Sections}

\newcommand{\mynum}[1]{{\fontsize{13}{13}\selectfont {#1}}}

\newcommand{\tablenum}[1]{{\fontsize{12}{13}\selectfont {#1}}}

\title{Token-Level Precise Attack on RAG: Searching for the Best Alternatives to Mislead Generation}

\author{
 \textbf{Zizhong Li\textsuperscript{1}} \quad
 \textbf{Haopeng Zhang\textsuperscript{2}} \quad
 \textbf{Jiawei Zhang\textsuperscript{1}} \quad
\\
 \textsuperscript{1}University of California, Davis \quad
 \textsuperscript{2}University of Hawaii at Mānoa
\\
   {\tt\small \{zzoli, jiwzhang\}@ucdavis.edu} \quad
   {\tt\small \{haopengz\}@hawaii.edu}
}

\begin{document}
\maketitle
\begin{abstract}
    While large language models (LLMs)
    have achieved remarkable success in providing trustworthy responses for knowledge-intensive tasks, they 
    still face critical limitations such as hallucinations and outdated knowledge. 
    To address these issues, the retrieval-augmented generation (RAG) framework enhances LLMs with access to external knowledge via a retriever, enabling more accurate and real-time outputs about the latest events.
    However, this integration brings new security vulnerabilities: the risk that malicious content in the external database can be retrieved and used to manipulate model outputs.
    Although prior work has explored attacks on RAG systems, existing approaches either rely heavily on access to the retriever or fail to jointly consider both retrieval and generation stages, limiting their effectiveness, particularly in black-box scenarios.
    To overcome these limitations, we propose Token-level Precise Attack on the RAG (TPARAG), a novel framework that targets both white-box and black-box RAG systems. TPARAG leverages a lightweight white-box LLM as an attacker to generate and iteratively optimize malicious passages at the token level, ensuring both retrievability and high attack success in generation.
    Extensive experiments on open-domain QA datasets demonstrate that TPARAG consistently outperforms previous approaches in retrieval-stage and end-to-end attack effectiveness. These results further reveal critical vulnerabilities in RAG pipelines and offer new insights into improving their robustness.
\end{abstract}

\section{Introduction}
The rapid advancement of large language models (LLMs) has led to impressive performance across a broad range of NLP tasks \cite{ouyang2022training, chang2024survey, guo2025deepseek}, including but not limited to knowledge-intensive generation \cite{singhal2025toward, wang2023knowledge, zhang2023summit, zhang2023extractive}.
However, LLM-generated content still faces challenges such as hallucinations and outdated knowledge \cite{liu2024survey, ji2023survey}. 
To address these limitations, the Retrieval-Augmented Generation (RAG) framework was introduced and has been widely adopted \cite{lewis2020retrieval, jiang2023active, chen2024benchmarking}.
RAG combines two core components: (1) a retriever that searches for relevant information from an external knowledge base, and (2) a reader that generates more accurate and informative responses based on the retrieved passages.

\begin{figure}
    \centering
    \includegraphics[width=1.0\linewidth]{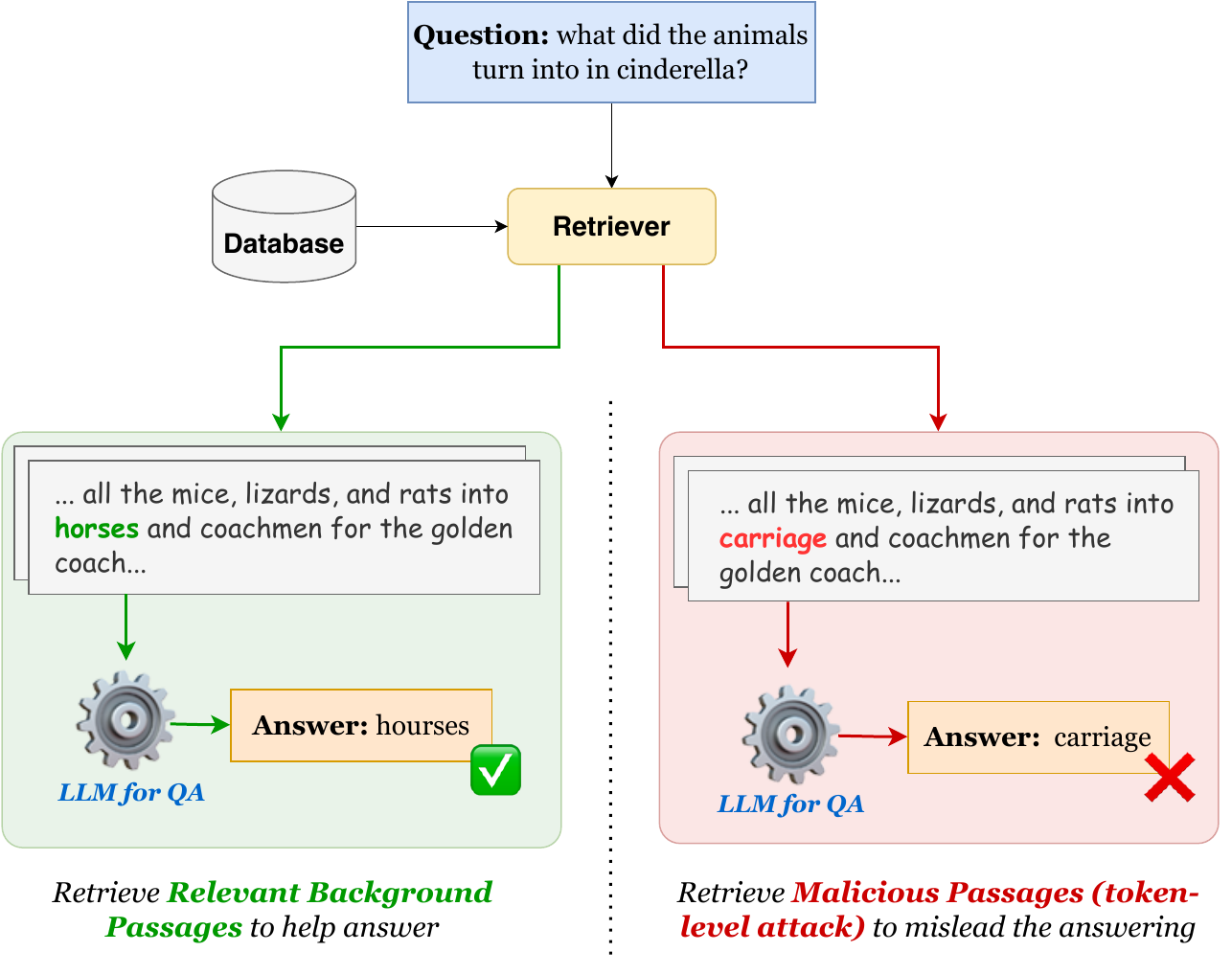}
    \caption{Comparison between the general RAG system (green background) and the RAG system attack (red background). The attacker replaces key information in the background knowledge to craft malicious passages, tricking the reader into generating an incorrect answer.}
    \label{fig:01}
    \vspace{-4mm}
\end{figure}

Although RAG enhances the generation quality of LLMs, it also introduces new potential risks to the reader's generation process, as the Figure \ref{fig:01} shows.
Since the retriever gathers information from external knowledge sources, the system is vulnerable to retrieving harmful or misleading content, which can cause the reader to generate incorrect or unsafe responses \cite{zeng2024good, jiang2024rag}. To exploit this weakness, recent studies have proposed attack methods that inject harmful information into the external database to manipulate the reader’s output \cite{zou2024poisonedrag, cheng2024trojanrag, xue2024badrag, cho2024typos}.

However, existing RAG attack approaches still face some of the following limitations:
1) Lack of joint consideration for the retrieval and generation. 
Prior studies \cite{zou2024poisonedrag, cheng2024trojanrag} often rely on teacher LLMs to generate malicious passages targeting the reader, but overlook whether the retriever can effectively retrieve these passages;
2) Overreliance on retriever models.
For instance, \citet{xue2024badrag} depends heavily on access to the retriever to generate query-similar malicious passages, which limits applicability in realistic black-box RAG settings; and
3) Lack of targeted control over reader outputs. The crafted malicious passage by the existing approaches is relatively random and cannot \textit{intentionally} guide the reader's responses \cite{cho2024typos, chen2024black}.

To address the limitations of existing RAG attack methods, we propose \textit{Token-level Precise Attack on RAG (TPARAG)}, a novel framework designed to target both black-box and white-box RAG systems.
TPARAG leverages a lightweight white-box LLM as the attacker to first generate malicious passages and then optimize them at the token level, ensuring that the adversarial content can be retrieved by the retriever and effectively misleads the reader.
Specifically, TPARAG operates in two stages: a generation attack stage and an optimization attack stage. In the generation stage, TPARAG records the top-\textit{k} most probable tokens at each position as potential substitution token candidates. 
These token candidates are then systematically recombined during optimization to construct a pool of malicious passage candidates.
Each passage candidate is further evaluated using two criteria to exploit vulnerabilities in both the retrieval and generation processes of the RAG pipeline: (1) the likelihood that the LLM attacker generates an incorrect answer given the passage, and (2) its textual similarity to the query.
The most effective malicious passage will finally be selected and injected into the external knowledge base to execute a targeted attack.

We conduct extensive experiments on multiple open-domain QA datasets using various lightweight white-box LLMs as attackers, under both black-box and white-box RAG settings.
The results demonstrate that TPARAG significantly outperforms prior approaches in attacking RAG systems, both in retrieval and end-to-end performance.
Moreover, we perform a series of quantitative analyses to identify key factors influencing attack success, critical vulnerabilities in RAG architectures and offering insights into improving their robustness.
In summary, our contributions are as follows.
\begin{itemize}
    \item We propose TPARAG, a token-level precise attack framework that leverages a lightweight white-box LLM as an attacker to exploit vulnerabilities in both the retrieval and generation stages of RAG systems.
    \item We conduct extensive experiments on multiple open-domain QA datasets, demonstrating the effectiveness of TPARAG in both black-box and white-box RAG settings.
    \item We perform detailed quantitative analyses to identify key factors influencing token-level attacks and explore strategies for maximizing attack effectiveness with minimal computational resources in practical scenarios.
\end{itemize}

\section{Related Work}

\subsection{Retrieval-augmented Generation}
Owing to the flexibility and effectiveness of the retriever model, it has become increasingly important to enhance LLM performance on knowledge-intensive tasks, driving the development of the RAG framework \cite{lewis2020retrieval, jiang2023active, gao2023retrieval}.
By incorporating information retrieval, RAG equips language models with more accurate and up-to-date background knowledge, improving generation quality and addressing common issues such as hallucination \cite{shuster2021retrieval, bechard2024reducing}.
Leveraging these advantages, RAG has been widely adopted across various NLP tasks, including but not limited to open-domain question answering \cite{siriwardhana2023improving, setty2024improving, wiratunga2024cbr}, summarization \cite{liu2024towards, suresh2024towards, edge2024local}, and dialogue systems \cite{vakayil2024rag, wang2024unims, kulkarni2024reinforcement}.

\begin{figure*}
    \centering
    \includegraphics[width=1.0\linewidth]{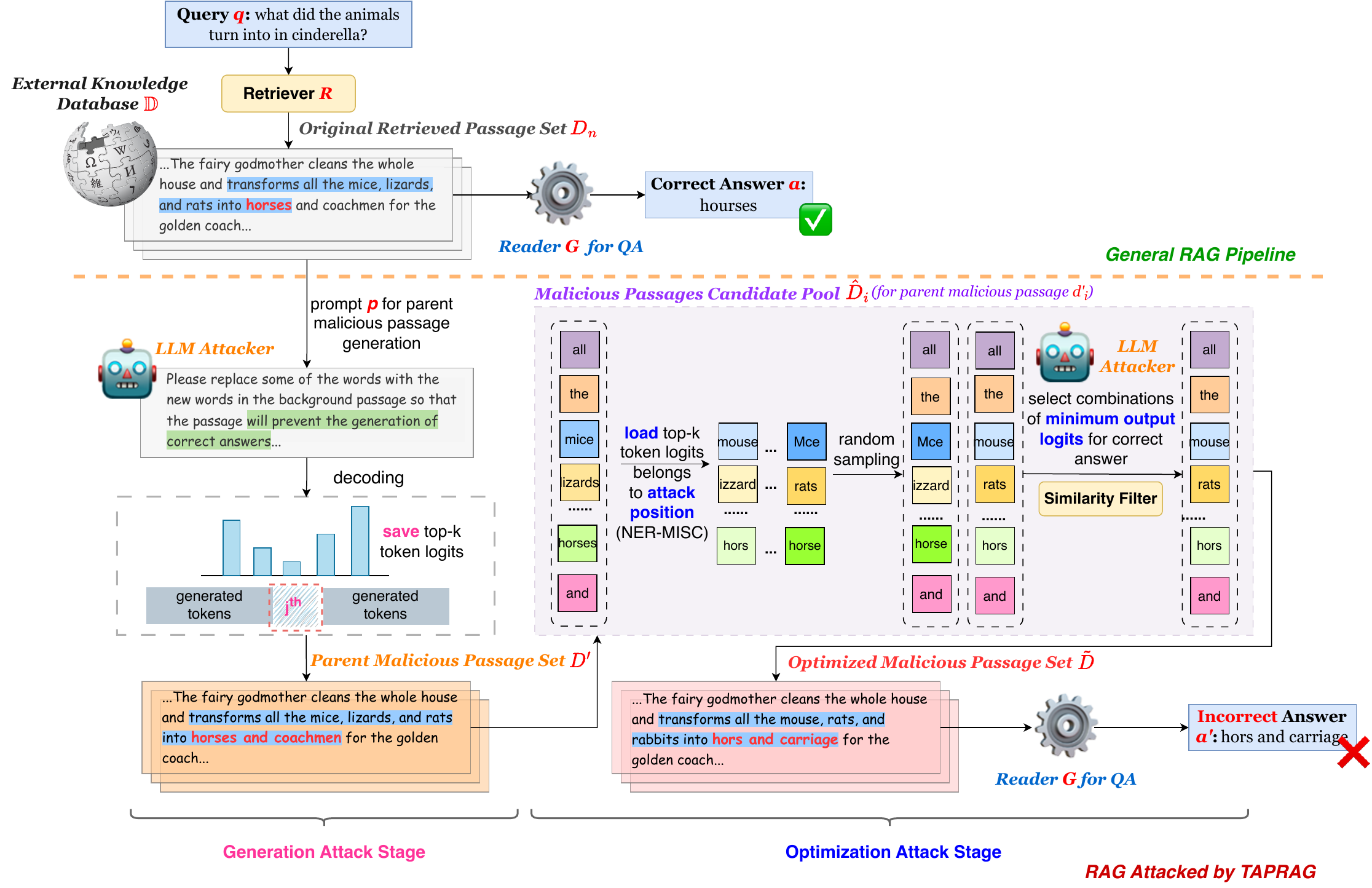}
    \caption{The framework of our proposed TPARAG. TPARAG first generates parent malicious passages through the generation attack stage (left). These passages are then recombined and refined during the optimization attack stage (right), producing optimized malicious passages that effectively mislead RAG's answer.}
    \label{fig:02}
    \vspace{-4mm}
\end{figure*}

\subsection{Risks in the RAG System}
While RAG generally produces more reliable and accurate outputs than only using LLMs,
its reliability remains subject to critical vulnerabilities.
Prior work \cite{zhou2024trustworthiness, ni2025towards, zeng2024good} have shown that the data leakage from the external retrieval database can significantly affect RAG’s reliability, primarily through two mechanisms: (1) the accurate extraction of sensitive or harmful information by the retriever, and (2) the generation of responses that expose such retrieved content.
As a result, the external dataset becomes a central point of attack.

Building on this vulnerability, previous RAG attack studies have focused on \textit{knowledge corruption}—injecting malicious content into the external database to mislead the readers' response \cite{wang2024astute, roychowdhury2024confusedpilot, zou2024poisonedrag, cheng2024trojanrag}.
For example, \citet{zou2024poisonedrag} and \citet{cheng2024trojanrag} use a teacher LLM to craft query-specific malicious content that guides the reader to generate incorrect outputs. 
However, these approaches often overlook whether the malicious content can actually be retrieved, posing a challenge in real-world scenarios where successful retrieval depends on sufficient similarity to the query.
To improve the recall rate of the crafted malicious content, later studies have introduced optimization techniques that align the malicious content with query semantics using embedding similarity scores provided by the retriever \cite{xue2024badrag, cho2024typos, jiang2024rag}. 
However, these approaches heavily rely on the retriever's parameters, limiting their applicability in black-box RAG systems.
Moreover, the optimization process often introduces excessive randomness, reducing control over the reader’s final output.
To address these limitations, we propose TPARAG, a token-level precise attack framework designed explicitly for black-box RAG settings. 
Additionally, TPARAG balances the likelihood of being retrieved with the ability to induce specific, incorrect responses from the reader.

\section{Proposed Method}
\label{sec:3}

\subsection{Problem Formulation}

\textbf{The Pipeline of RAG.} A typical RAG system consists of two main components: a retriever model $R$ (parameterized with $\theta_R$) that retrieves relevant information from an external database, and a reader $G$ (parameterized with $\theta_G$) that generates responses based on the retrieved content.
Specifically, given a query $q$, the goal of the retriever model $R$ is to find a subset of the most relevant passages $D=\{d_1, d_2, ..., d_m\}$ from the knowledge database $\mathbb{D}$, where each $d_i$ represents a unique passage.
Then, this subset $D$ is combined with the query $q$ to form the prompt input of the reader $G$, and then generates the corresponding answer $a$ as follows:
\begin{align}
    a = G(q, D; \theta_G),
\end{align}
\textbf{RAG Attack's Objective.} 
Given a query $q$, the objective of the RAG attack framework is to construct a malicious passage subset $\tilde{D} = \{\tilde{d}_1, \tilde{d}_2, ..., \tilde{d}_m\}$ and inject it into the knowledge database $\mathbb{D}$.
The goal is for the retriever $R$ to select $\tilde{d}_i \in \tilde{D}$ and forward it to the reader $G$, thereby inducing the generation for an incorrect response $a'$:
\begin{align}
    a' = G(q, \tilde{D}; \theta_G),
\end{align}
The construction of each $\tilde{d}_i \in \tilde{D}$ focuses on two key objectives: 1) the retrieval attack, and 2) the generation attack.
The first one aims to prioritize the retrieval of $\tilde{d}_i \in \tilde{D}$ over $d_i \in D$. That is, maximizing the textual similarity between the given query $q$ and each malicious passage $\tilde{d}_i$:
\begin{align}
    \label{equ:03}
    \max {s(q, \tilde{d}_i; \theta_R)} \quad \text{s.t.} \frac{s(q, \tilde{d}_i; \theta_R)}{s(q, d_i; \theta_R)} > 1,
\end{align}
where $s(\cdot; \theta_R)$ denotes the cosine similarity between the embedding of the query $q$ and the malicious passage $\tilde{d}_i$ encoded by the retriever $R$.

For the generation attack, the objective is to reduce the likelihood that incorporating $\tilde{d}_i$ into the reader leads to the correct answer $a$, compared to $d_i$, and seeks to minimize this likelihood to the greatest extent possible:
\begin{align}
    \label{equ:04}
    \min {P_{G}(a|q, \tilde{d}_i; \theta_G)} \quad \text{s.t.} \frac{P_{G}(a|q, \tilde{d}_i; \theta_G)}{P_{G}(a|q, d_i; \theta_G)} < 1,
\end{align}
where $P_{G}(a|\cdot ; \theta_G)$ denotes the likelihood that the reader $G$ generates the correct answer $a$.

\subsection{Token-level Precise Attack on RAG }
To achieve a dynamic balance between Equation (\ref{equ:03}) and (\ref{equ:04}), we design TPARAG, which leverages lightweight white-box LLMs to generate and then optimize the malicious passage subset $\tilde{D}$ at the token-level, enabling precise, query-specific attacks on the RAG systems.

As shown in Figure \ref{fig:02}, given a query $q$, TPARAG first uses a white-box LLM attacker $LLM_{attack}$ to fabricate an incorrect answer and generate a corresponding malicious background passage $\tilde{d}_i$ based on $d_i$, while maintaining high textual similarity with $q$.
During this generating process, the attacker model $LLM_{attack}$ records the top-\textit{k} most probable generated tokens at each position. 
These tokens are then recombined in the optimization to form new malicious candidate passages at key positions by modifying key token positions.
Then, the candidates are further filtered based on 1) their likelihood of inducing incorrect answers; and 2) their textual similarity to the query $q$. The final optimized malicious passage set $\tilde{D}$ is then selected from this refined pool.
The following subsections detail each stage of the TPARAG pipeline, and we also provide the detailed algorithm in Appendix \ref{appendix}.

\subsubsection{Initialization}
\textbf{Threshold for Generation Attack.} TPARAG begins with a data initialization step.
Given a query $q$ and a set of $m$ originally retrieved relevant background passages $D=\{d_1, d_2, ..., d_m\}$, TPARAG establishes a threshold $l$ that filters out less effective malicious passages, retaining only those more likely to mislead the reader:
\begin{equation}
    l = \max_{l_i \in L} l_i,
\end{equation} 
where $L=\{l_1, l_2, ..., l_m\}$ is the generation logits for the correct answer $a$ using the output of the white-box LLM attacker $LLM_{attack}$. 
\\
\textbf{Threshold for Retrieval Attack.} TPARAG computes textual similarity scores $S=\{s_1, s_2, ..., s_m\}$ between the query $q$ and each passage $d_i$, using them to define a similarity threshold $s$:
\begin{equation}
    s=\min_{s_i \in S} s_i,
\end{equation}
This threshold helps identify malicious passages that are more likely to be retrieved by the retriever. \\ 
\textbf{Entity-Based Attack Localization.} TPARAG employs a named entity recognition (NER) tool \cite{akbik2019flair} as the attack locator $Loc$ to annotate the correct answer $a$ with entity labels:
\begin{equation}
    pos_{attack} = Loc(a),
\end{equation}
which helps to identify the specific token types to be targeted in its optimization attack stages.

\subsubsection{Generation Attack Stage}
\begin{figure}[h]
    \centering
    \includegraphics[width=1.0\linewidth]{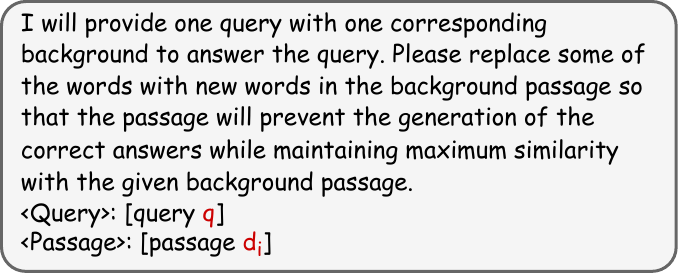}
    \caption{An example of the prompt for TPARAG's generation attack stage.}
    \label{fig:03}
    \vspace{-4mm}
\end{figure}

After the initialization, TPARAG uses the white-box $LLM_{attack}$ to generate a set of parent malicious passages subset $D'=\{d'_1, d'_2, ..., d'_m\}$, based on the query $q$ and each background passage in $D$, following a predefined prompt $p$ as Figure \ref{fig:03} shows.
Therefore, the attacker generates parent malicious passages as:
\begin{equation}
    d'_i=LLM_{attack}(q, d_i; \theta_{LLM_{attack}}),
\end{equation}
During the decoding process, the LLM attacker records, at each token position $j$, the top-$k$ tokens $T_{ij}$ with the highest generation probabilities:
\begin{equation}
    T_{ij} =\mbox{Top-}k(P_{LLM_{attack}}(t_{1:j-1}, p)),
\end{equation}
where $t_{1:j-1}$ denotes the partially generated sequence up to position $j-1$ and $p$ denotes the input prompt.
These top-\textit{k} tokens are later used in the optimization attack stage to further refine the parent malicious passages.

\subsubsection{Optimization Attack Stage}
\textbf{Entity-Based Token Filtering.} In this stage, TPARAG first reuses the attack locator $Loc$ to perform NER on each token position $j$ in $d'_i$, identifying tokens that share the same entity type as the target position $pos_{attack}$. 
These tokens are then used as candidate substitution points for the token-level attack. \\
\textbf{Token Substitution and Candidate Generation.} 
For each parent malicious passage $d'_i \in D'$, TPARAG randomly selects the tokens that share the same entity type as the target position $pos_{attack}$, based on a predefined maximum token substitution rate $pr_{sub}$ (i.e., the upper bound on the proportion of tokens to be replaced). 
For each position $j$ selected for substitution, TPARAG generates multiple variants by replacing the token with alternatives from the previously recorded top-\textit{k} token set $T_{ij}$.
This process produces a set of optimized malicious passage candidates for each $d'_i$, denoted as:
\begin{equation}
    \hat{D}_i = \{\hat{d}_{i1}, \hat{d}_{i2}, ..., \hat{d}_{in}\},
\end{equation}
where $n$ is the number of generated candidates. \\
\textbf{Similarity-Based Filtering.} For each candidate $\hat{d}_{ij} \in \hat{D}_i$, TPARAG computes its textual similarity to the query $q$ and compares it with the predefined similarity threshold $s$. Candidates with similarity below the threshold are discarded. 

TPARAG supports attacks under both white-box and black-box RAG settings. 
In the white-box setting, the similarity between each candidate passage and the query is computed using embeddings from the original retriever. 
In the black-box setting, where access to the retriever is unavailable, TPARAG uses Sentence-BERT \cite{reimers2019sentence} as a substitute to estimate the textual similarity, which also proves  to be highly effective in the attack process. \\
\textbf{Likelihood-Based Selection.} For the remaining passage candidates generated from $d'_i$, the LLM attacker $LLM_{attack}$ simulates the generation process of the RAG system by computing the likelihood $\hat{l}_{ik}$ of generating the correct answer $a$, given the query $q$ and candidate passage $\hat{d}_{ik}$:
\begin{equation}
    \hat{l}_{ik} = P_{LLM_{attack}}(a|q, \hat{d}_{ik}; \theta_{LLM_{attack}}).
\end{equation}
Among all candidates in $\hat{D}_i$, the one with the lowest $\hat{l}_{ik}$ is selected as the final optimized malicious passage $\tilde{d}_i$.

Moreover, TPARAG performs iterative optimization over multiple rounds of malicious passage generation. 
In each iteration, it applies this greedy search strategy to select the candidate with the strongest attack effect like $\tilde{d}_i$, which is then used as input for the next round. 

Following this strategy, the resulting set of optimized malicious passages $\tilde{D}$ maintains high textual similarity with the query $q$, while introducing precise and effective interference to disrupt the generation of the correct answer.

\section{Experiment}
\label{sec:04}
\begin{table*}[htbp]
    \centering
    {\renewcommand{\arraystretch}{1.2}
    \resizebox{1.05\textwidth}{!}{%
    \begin{tabular}{ 
    c|c|  
    *{5}{>{\centering\arraybackslash}p{1.25cm}}|  
    *{5}{>{\centering\arraybackslash}p{1.25cm}}|  
    *{5}{>{\centering\arraybackslash}p{1.25cm}} 
    }
    \toprule
    \toprule
    \multirow{3}{*}{\textbf{RAG Setting}} & \multirow{3}{*}{\textbf{LLM Attacker}} 
    & \multicolumn{5}{c|}{\textbf{NQ}} 
    & \multicolumn{5}{c|}{\textbf{TriviaQA}} 
    & \multicolumn{5}{c}{\textbf{PopQA}} \\
    \cmidrule(lr){3-7} \cmidrule(lr){8-12} \cmidrule(lr){13-17}
    & & \bm{$ASR_R \uparrow$} & \bm{$ASR_L\uparrow$} & \bm{$ASR_T \uparrow$} & \textbf{  EM$\downarrow$} & \textbf{F1$\downarrow$} 
    & \bm{$ASR_R \uparrow$} & \bm{$ASR_L \uparrow$} & \bm{$ASR_T \uparrow$} & \textbf{  EM$\downarrow$} & \textbf{F1 $\downarrow$}
      & \bm{$ASR_R \uparrow$} & \bm{$ASR_L \uparrow$} & \bm{$ASR_T \uparrow$} & \textbf{  EM$\downarrow$} & \textbf{F1$\downarrow$} \\
    \midrule
    \midrule
    \multirow{4}{*}{\mynum{Black-box (TPARAG)}} 
    & QWen2.5-3B  & \mynum{77.2} & \mynum{99.6} & \mynum{77.2} & \mynum{68.0} & \mynum{76.2} & \mynum{90.1} & \mynum{74.2} & \mynum{67.1} & \mynum{78.0} & \mynum{80.8} & \mynum{84.6} & \mynum{88.6} & \mynum{75.6} & \mynum{48.0} & \mynum{50.1} \\
    & QWen2.5-7B  & \mynum{79.0} & \mynum{\textbf{100.0}} & \mynum{\textbf{79.0}} & \mynum{65.0} & \mynum{73.2} & \mynum{85.6} & \mynum{77.2} & \mynum{65.1} & \mynum{\textbf{56.0}} & \mynum{\textbf{60.4}} & \mynum{85.8} & \mynum{85.8} & \mynum{73.4} & \mynum{28.0} & \mynum{34.8} \\
    & Mistral-7B  & \mynum{76.7} & \mynum{\textbf{100.0}} & \mynum{76.7} & \mynum{\textbf{45.0}} & \mynum{\textbf{55.2}} & \mynum{\textbf{94.8}} & \mynum{67.4} & \mynum{64.5} & \mynum{58.0} & \mynum{61.5} & \mynum{92.2} & \mynum{87.2} & \mynum{81.2} & \mynum{\textbf{7.0}} & \mynum{\textbf{9.5}} \\
    & Gemma2-9B & \mynum{66.4} & \mynum{98.0} & \mynum{65.6} & \mynum{55.0} & \mynum{62.9} & \mynum{73.4} & \mynum{\textbf{98.6}} & \mynum{\textbf{72.6}} & \mynum{68.0} & \mynum{72.9} & \mynum{76.4} & \mynum{91.8} & \mynum{69.8} & \mynum{32.0} & \mynum{34.6} \\
    \midrule
    \multirow{4}{*}{\mynum{Black-box (PoisonedRAG)}} 
    & QWen2.5-3B  & \mynum{48.0} & \mynum{46.2} & \mynum{39.8} & \mynum{66.0} & \mynum{57.0} & \mynum{92.8} & \mynum{64.0} & \mynum{59.5} & \mynum{77.0} & \mynum{79.2} & \mynum{96.2} & \mynum{77.1} & \mynum{73.8} & \mynum{42.0} & \mynum{45.3} \\
    & QWen2.5-7B & \mynum{39.2} & \mynum{54.0} & \mynum{19.8} & \mynum{74.0} & \mynum{80.7} & \mynum{22.0} & \mynum{59.6} & \mynum{13.9} & \mynum{91.0} & \mynum{93.0} & \mynum{78.8} & \mynum{66.3} & \mynum{53.7} & \mynum{36.0} & \mynum{39.4} \\
    & Mistral-7B & \mynum{\textbf{81.5}} & \mynum{72.8} & \mynum{57.6} & \mynum{72.0} & \mynum{76.4} & \mynum{91.3} & \mynum{76.1} & \mynum{70.7} & \mynum{77.0} & \mynum{82.2} & \mynum{94.8} & \mynum{73.2} & \mynum{68.0} & \mynum{38.0} & \mynum{43.3} \\
    & Gemma2-9B & \mynum{67.6} & \mynum{86.2} & \mynum{59.0} & \mynum{50.0} & \mynum{56.1} & \mynum{90.6} & \mynum{70.0} & \mynum{64.6} & \mynum{69.0} & \mynum{73.2} & \mynum{\textbf{97.8}} & \mynum{\textbf{92.9}} & \mynum{\textbf{90.7}} & \mynum{24.0} & \mynum{28.9} \\
    \midrule
    \midrule
    \multirow{4}{*}{\mynum{White-box (TPARAG)}}
    & QWen2.5-3B  & \mynum{83.6} & \mynum{87.2} & \mynum{72.4} & \mynum{59.0} & \mynum{65.7} & \mynum{99.8} & \mynum{70.3} & \mynum{70.2} & \mynum{64.0} & \mynum{68.6} & \mynum{99.6} & \mynum{89.6} & \mynum{89.2} & \mynum{39.0} & \mynum{45.0} \\
    & QWen2.5-7B  & \mynum{99.4} & \mynum{99.4} & \mynum{99.2} & \mynum{71.0} & \mynum{75.6} & \mynum{\textbf{100.0}} & \mynum{\textbf{85.6}} & \mynum{\textbf{85.6}} & \mynum{60.0} & \mynum{64.2} & \mynum{99.8} & \mynum{84.6} & \mynum{84.4} & \mynum{15.0} & \mynum{17.5} \\
    & Mistral-7B & \mynum{99.4} & \mynum{99.2} & \mynum{98.6} & \mynum{52.0} & \mynum{\textbf{61.7}} & \mynum{\textbf{100.0}} & \mynum{69.1} & \mynum{69.1} & \mynum{\textbf{52.0}} & \mynum{\textbf{57.9}} & \mynum{\textbf{100.0}} & \mynum{85.0} & \mynum{85.0} & \mynum{\textbf{9.0}} & \mynum{\textbf{10.3}} \\
    & Gemma2-9B & \mynum{\textbf{100.0}} & \mynum{\textbf{99.8}} & \mynum{\textbf{99.8}} & \mynum{\textbf{50.0}} & \mynum{62.7} & \mynum{\textbf{100.0}} & \mynum{74.9} & \mynum{74.9} & \mynum{77.0} & \mynum{80.9} & \mynum{\textbf{100.0}} & \mynum{\textbf{91.2}} & \mynum{\textbf{91.2}} & \mynum{33.0} & \mynum{33.5} \\
    \midrule
    \mynum{White-box (GARAG)*} & Mistral-7B & \mynum{87.5} & \mynum{85.5} & \mynum{73.3} & \mynum{63.9} & \mynum{--} & \mynum{88.8} & \mynum{86.4} & \mynum{75.2} & \mynum{66.2} & -- & -- & -- & -- & -- & -- \\
    \midrule
    \multicolumn{2}{c|}{\mynum{White-box (PoisonedRAG)}} & \mynum{\textbf{100.0}} & \mynum{89.8} & \mynum{89.8} & \mynum{83.0} & \mynum{87.9} & \mynum{\textbf{100.0}} & \mynum{66.3} & \mynum{66.3} & \mynum{94.0} & \mynum{96.9} & \mynum{\textbf{100.0}} & \mynum{71.8} & \mynum{71.8} & \mynum{87.0} & \mynum{93.1} \\
    \midrule
    \midrule
    \multicolumn{2}{c|}{\mynum{w/o RAG}}  & \mynum{--} & \mynum{--} & \mynum{--} & \mynum{67.0} & \mynum{76.4} & \mynum{--} & \mynum{--} & \mynum{--} & \mynum{94.0} & \mynum{95.5} & \mynum{--} & \mynum{--} & \mynum{--} & \mynum{70.0} & \mynum{74.0} \\
    \midrule
    \multicolumn{2}{c|}{\mynum{RAG}}  & \mynum{--} & \mynum{--} & \mynum{--} & \mynum{82.0} & \mynum{88.0} & \mynum{--} & \mynum{--} & \mynum{--} & \mynum{97.0} & \mynum{98.5} & \mynum{--} & \mynum{--} & \mynum{--} & \mynum{92.0} & \mynum{96.3} \\
    \bottomrule
    \bottomrule
    \end{tabular}
    }
    \caption{Comparison of LLM attackers with different RAG settings across three QA datasets.}
    \label{tab:llm-retriever-comparison}
    \vspace{-3mm}
    }
\end{table*}
\subsection{Experiment Setup}
\textbf{Dataset.} We conduct experiments on three open-domain question-answering benchmark datasets: NaturalQuestions (NQ) \cite{kwiatkowski2019natural}, TriviaQA \cite{joshi2017triviaqa}, and PopQA \cite{mallen2022not}.
For the external knowledge database, we use Wikipedia data dated December 20, 2018, adapting the passage embeddings provided by ATLAS \cite{izacard2023atlas}. 
We randomly sample 100 query instances from each dataset's training set as the targeted queries for the TPARAG attack. \\
\textbf{Evaluation Metrics.} 
Following the previous work \cite{cho2024typos}, we decompose ASR into three components: $ASR_R$(\%), $ASR_L$(\%), and $ASR_T$(\%), which denote the attack success percentage of the retrieval, the attack success percentage of the generation, and the overall attack success percentage, respectively.
Specifically, $ASR_R$ measures the proportion of the crafted malicious passages satisfying Equation (\ref{equ:03}) (i.e., $\frac{s(q, \tilde{d}_i; \theta_R)}{s(q, d_i; \theta_R)} > 1$), $ASR_L$ measures the proportion of malicious passages satisfying Equation (\ref{equ:04}) (i.e., $\frac{P_{G}(a|q, \tilde{d}_i; \theta_G)}{P_{G}(a|q, d_i; \theta_G)} < 1$), and $ASR_T$ measures the proportion of malicious passages satisfying both conditions.

Meanwhile, we report the standard Exact Match (EM) and F1-Score, which show the accuracy and precision of the generated responses, to evaluate the end-to-end attack performance.\\
\textbf{Models.} For the RAG system, we choose the closed-source GPT-4o \cite{openai2023gpt4} as its reader, and the off-the-shelf Contriever \cite{izacard2021unsupervised} as the retriever model.
During the TPARAG attack, we leverage various lightweight white-box LLMs, including Qwen2.5-3B, Qwen2.5-7B \cite{qwen2025qwen25technicalreport}, Mistral-7B \cite{jiang2023mistral7b}, and Gemma2-9B \cite{gemmateam2024gemma2improvingopen}, as the LLM attackers.
\setlength{\tabcolsep}{1.5mm}{
\begin{table}[t]
    \centering
    \scalebox{1.0}{
    \resizebox{0.5\textwidth}{!}{
    \begin{tabular}{
        >{\centering\arraybackslash}p{2.5cm}  
        >{\centering\arraybackslash}p{2.2cm}  
        *{5}{>{\centering\arraybackslash}p{1.3cm}}  
    }
      \toprule
      \multirow{3}*{\textbf{Attack Setting}} & \multirow{3}*{\textbf{LLM Attacker}} & \multicolumn{5}{c}{\textbf{Evaluation Metrics}} \\
      \cmidrule(lr){3-7}
      & & \textbf{\bm{$ASR_R\uparrow$}} & \textbf{\bm{$ASR_L\uparrow$}} & \textbf{\bm{$ASR_T\uparrow$}} & \textbf{EM$\downarrow$} & \textbf{F1$\downarrow$} \\
      \midrule
       \multirow{4}*{Initialization} & QWen2.5-3B & \tablenum{77.2} & \tablenum{99.6} & \tablenum{77.2} & \tablenum{68.0} & \tablenum{76.2} \\
      & QWen2.5-7B & \tablenum{71.8} & \tablenum{100.0} & \tablenum{71.8} & \tablenum{66.0} & \tablenum{74.1} \\
      & Mistral-7B & \tablenum{76.7} & \tablenum{100.0} & \tablenum{76.7} & \tablenum{45.0} & \tablenum{55.2} \\
      & Gemma2-9B & \tablenum{66.4} & \tablenum{98.0} & \tablenum{65.6} & \tablenum{55.0} & \tablenum{62.9} \\
      \midrule
      \multirow{4}*{w/o Initialization} & QWen2.5-3B & \tablenum{74.0} & \tablenum{99.6} & \tablenum{73.6} & \tablenum{79.0} & \tablenum{83.6} \\
      & QWen2.5-7B & \tablenum{74.8} & \tablenum{100.0} & \tablenum{74.8} & \tablenum{63.0} & \tablenum{66.7} \\
      & Mistral-7B & \tablenum{70.2} & \tablenum{99.0} & \tablenum{69.6} & \tablenum{68.0} & \tablenum{73.2} \\
      & Gemma2-9B & \tablenum{65.7} & \tablenum{99.6} & \tablenum{65.7} & \tablenum{52.0} & \tablenum{59.0} \\
      \bottomrule
    \end{tabular}
    }}
    \caption{TPARAG’s performance on the NQ dataset under black-box RAG setting without initialization from the original relevant background passages.}
    \label{tab:performance-drp}
    \vspace{-5mm}
\end{table}}

We maintain the same LLM attacker in the generation and optimization stage. \\
\textbf{Baselines.} To ensure a fair and reproducible comparison, we choose two representative prior approaches as our experimental baseline -- PoisonedRAG \cite{zou2024poisonedrag} and GARAG \cite{cho2024typos}.
In the white-box setting, we replicate the PoisonedRAG setup using Contriever \cite{izacard2021unsupervised} as the publicly available retriever.
In addition, we conduct baseline evaluations using the original RAG system and a reader-only QA setup (i.e., w/o RAG), which serve as reference points for comparison. \\
\textbf{Implementation Details.} Considering the trade-off between computational cost and performance improvement, we set the maximum iteration number $N_{iter}$ to 5, the candidate malicious passage subset size to 20, and the maximum token substitution rate as $0.2$. 
Additionally, we restrict the size of the relevant document subset $D$ to 5, and each retrieved passage has a maximum length of 128.
We will further discuss the impact of different parameters on the attack outcomes in Section \ref{sec:05}.
\subsection{Experimental Results}
Table \ref{tab:llm-retriever-comparison}\footnote{Best performance values are highlighted in bold. Results marked with * are from the original paper; others are tested with our implementation and datasets.} presents the experimental results on selected attacked query instances from three datasets, comparing our proposed TPARAG framework with other baselines. 
For TPARAG, we report the end-to-end performance corresponding to the iteration with the highest $ASR_T$.
The results show that \textbf{TPARAG effectively performs end-to-end attacks while simultaneously increasing the likelihood of malicious passages retrieval across both black-box and white-box RAG settings}.

Specifically, in the black-box setting, TPARAG achieves a retrieval attack success rate exceeding 66\%, reaching up to 94\% in the best case. 
In the white-box setting, its performance improves further, reaching a 100\% success rate in half of the test cases and maintaining a minimum of 83\%, underscoring TPARAG's effectiveness in misleading the retriever regardless of system access level.
Meanwhile, TPARAG's optimized malicious passages demonstrate strong effectiveness for the attack in the generation process.
In the black-box setting, it achieves at least a 67\% attack success rate against the LLM attacker, indicating its ability to disrupt the answer generation without internal model access.
In the white-box setting, the minimum end-to-end attack success rate further rises to 70\%, indicating that TPARAG can more precisely target model vulnerabilities when partial access is available.
These results confirm that  \textbf{TPARAG's token-level passage construction is effective for retrieval and misleading answer generation}.

\begin{figure}
    \centering
\includegraphics[width=1.0\linewidth]{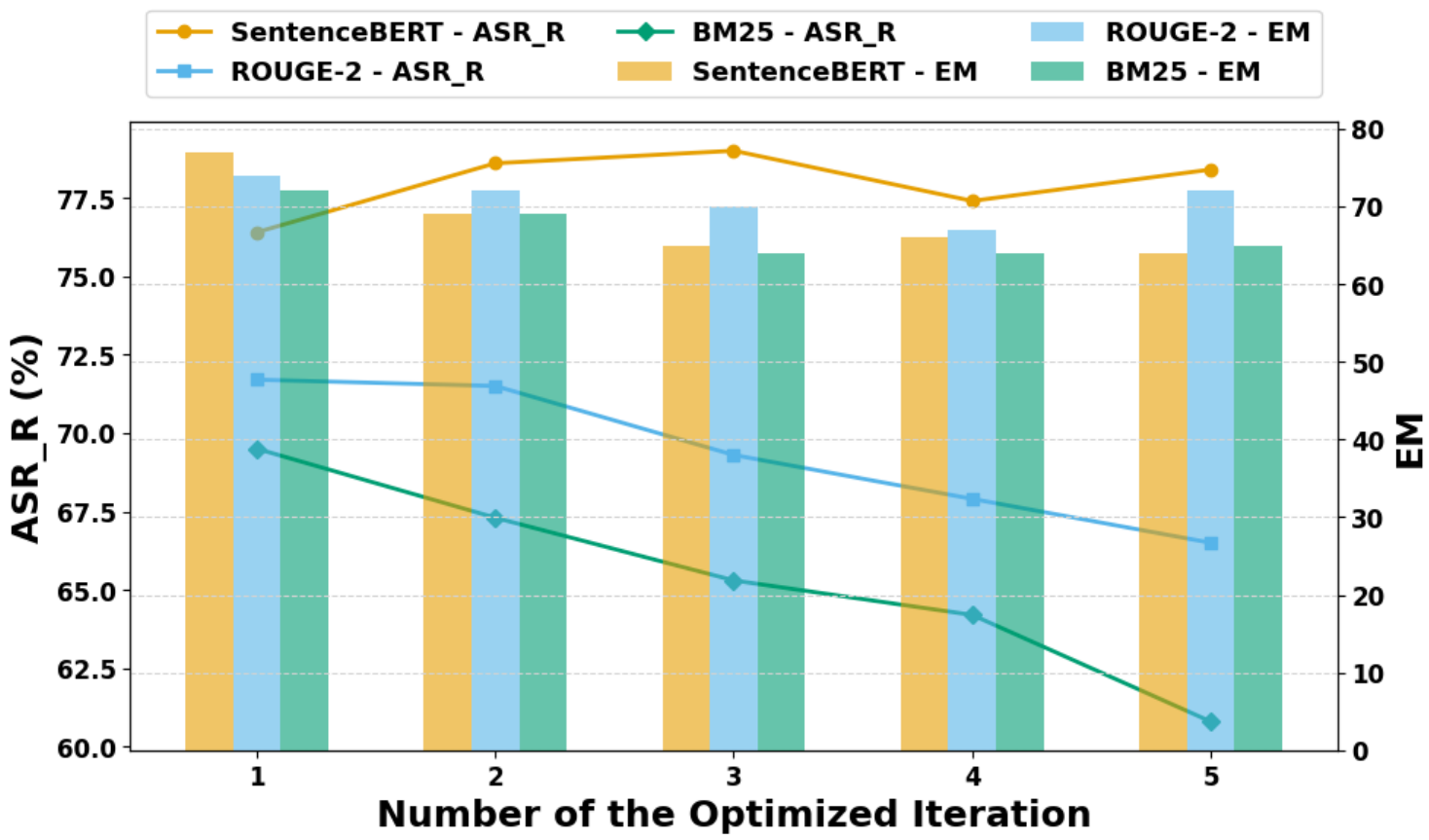}
    \caption{The impact of different similarity filters on TPARAG performance under black-box RAG setting.}
    \label{fig:analysis_sim}
    \vspace{-4mm}
\end{figure}

\begin{figure}
    \centering \includegraphics[width=1.0\linewidth]{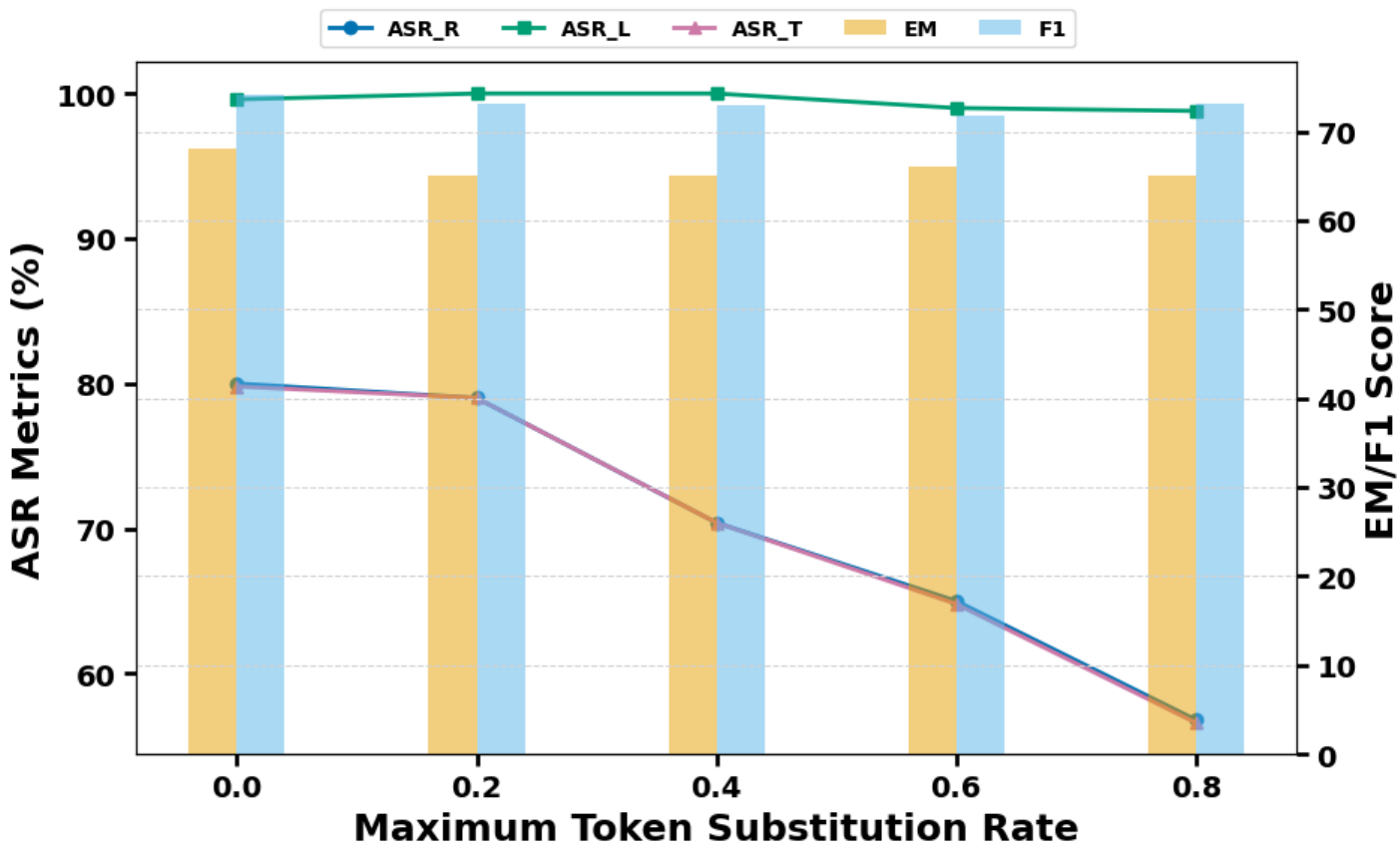}
    \caption{The performance of TPARAG under different maximum token substitution rates.}
    \label{fig:analysis_sec2}
    \vspace{-4mm}
\end{figure}

Regarding the end-to-end attack performance, \textbf{TPARAG consistently reduces the answer accuracy of the RAG systems} under both black-box and white-box settings.
In all cases, accuracy drops below that of the original RAG baseline, and in most instances, it even falls below the performance of using the reader alone without any retrieved context.
However, we observe that \textbf{the LLM attacker's success rate ($ASR_L$) does not always align with the end-to-end attack effectiveness}.
Two factors may cause this discrepancy: 1) limitations of $ASR_L$: while it reflects whether the attack successfully reduces the likelihood of generating the correct answer using malicious passages, it does not capture the \textit{degree} of reduction or its actual impact on final answer quality; 2) model discrepancy: behavioral and sensitivity differences between the white-box LLM attacker and the black-box reader may lead to misalignment during generation.

\textbf{Compared to the PoisonedRAG and GARAG baselines, TPARAG offers more robust and consistent performance}. Under the white-box RAG setting, experimental results show that GARAG underperforms TPARAG in both retrieval and end-to-end attack effectiveness when using the same LLM attacker.
In addition, although PoisonedRAG achieves perfect $ASR_R$ under the white-box RAG setting via gradient-based optimization, it is considerably less effective at attacking the generation stage. 
In the black-box RAG setting, PoisonedRAG also suffers from unstable attack performance (i.e., $ASR_L$ varies significantly depending on the choice of LLM attacker), and it underperforms TPARAG in most end-to-end evaluations.

\section{Analysis}
\label{sec:05}
\subsection{Strategy for Constructing Malicious Passages in Black-box Generation Attack}

Under the black-box RAG setting, our proposed TPARAG initializes the malicious passage using the original relevant background passages $D$.
However, this step still introduces a degree of dependency on the retriever within the RAG system.
Therefore, we further evaluate TPARAG's performance in a \textit{fully black-box RAG setting}, where the generation attack stage is conducted without any initialization from $D$. Instead, the LLM attacker directly generates malicious passages based solely on the query, followed by token-level optimization using the TPARAG framework.

As shown in the Table \ref{tab:performance-drp}, initializing with the original background passages generally leads to better retrieval-stage performance (i.e., higher $ASR_R$), suggesting that mimicking original content helps the malicious passages better deceive the retriever.
However, even without such initialization, TPARAG still achieves strong end-to-end attack performance: both EM and F1-Score drop significantly compared to the RAG baseline, demonstrating the feasibility and effectiveness of TPARAG in a fully black-box RAG setting.

\subsection{Different Similarity Signal in Black-box Optimization Attack}

As described in Section \ref{sec:3}, TPARAG uses Sentence-BERT as a substitute for the retriever to implement the similarity filtering mechanism when attacking black-box RAG systems. 
In this subsection, we further investigate the effectiveness of Sentence-BERT as an alternative similarity signal in TPARAG by comparing it with two widely used text similarity metrics: ROUGE-2 \cite{lin2004rouge} and BM25 \cite{robertson2009probabilistic}, while keeping all other experimental settings unchanged.
As shown in Figure \ref{fig:analysis_sim}, Sentence-BERT most effectively approximates the behavior of BERT-based retrievers. It enables TPARAG to iteratively optimize malicious passages, leading to an increasing trend in $ASR_R$ over successive optimization steps, while maintaining strong end-to-end attack performance.
\begin{figure}
    \centering
    \includegraphics[width=1.0\linewidth]{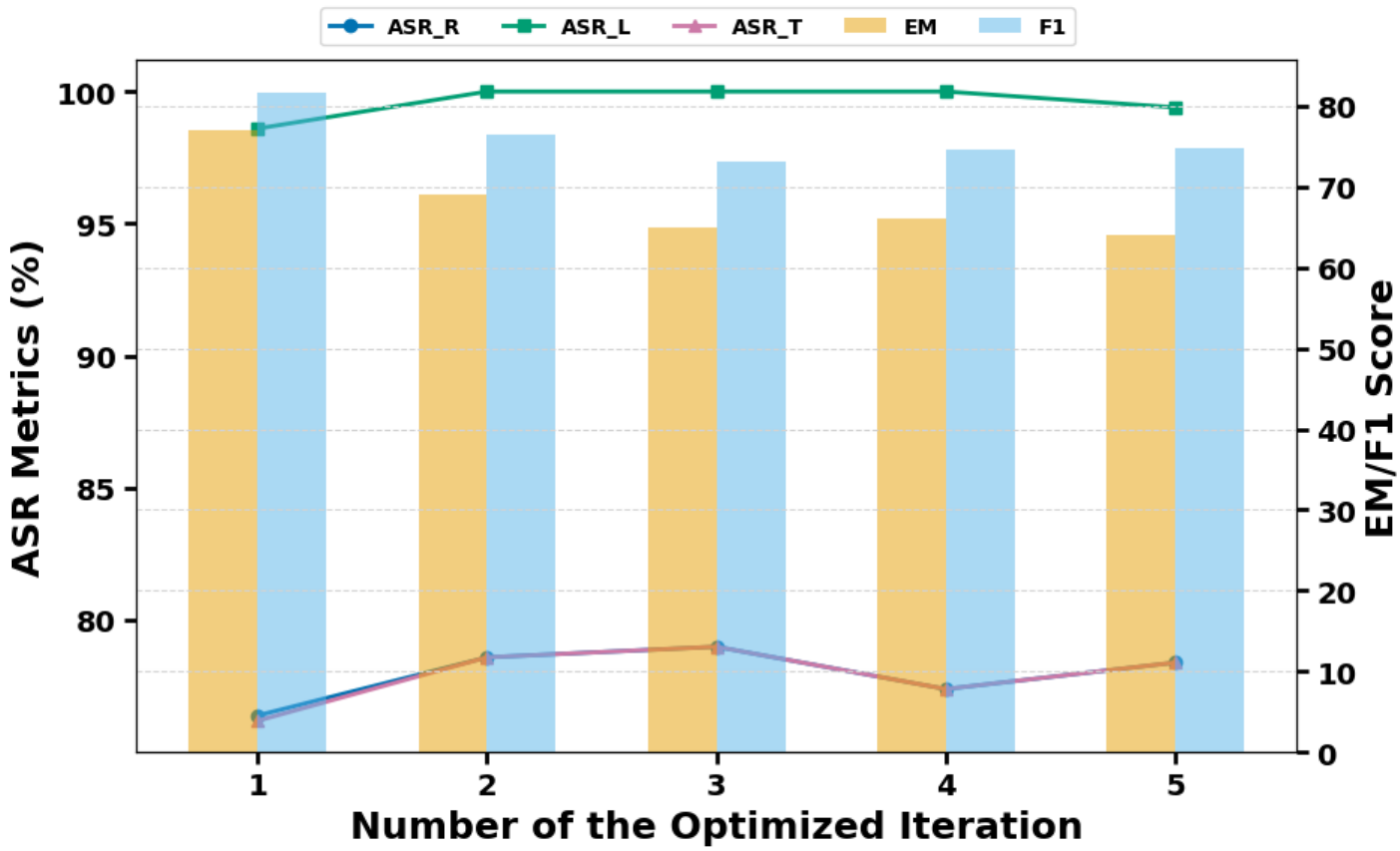}
    \caption{The performance variation of TPARAG across different numbers of optimization iterations.}
    \label{fig:analysis_iter}
    \vspace{-2mm}
\end{figure}
In contrast, ROUGE-2 or BM25 fails to enhance the retrieval attack effectiveness. Furthermore, when ROUGE-2 is used, TPARAG shows inconsistent end-to-end attack success, highlighting its limitations as a similarity proxy in this context.

\subsection{Impact of the Hyperparameter}
\textbf{Maximum Token Substitution Rate.} To evaluate the impact of the maximum token substitution rate $pr_{sub}$ on TPARAG’s performance, we very the $pr_{sub}$ (0.0, 0.2, 0.4, 0.6, and 0.8) under the black-box RAG while keeping all other settings fixed.
As shown in Figure \ref{fig:analysis_sec2}, higher substitution rates allow greater flexibility in modifying the malicious passages but also increase semantic divergence from the original context, which in turn reduces their likelihood of being retrieved.
\begin{figure}
    \centering
    \includegraphics[width=1.0\linewidth]{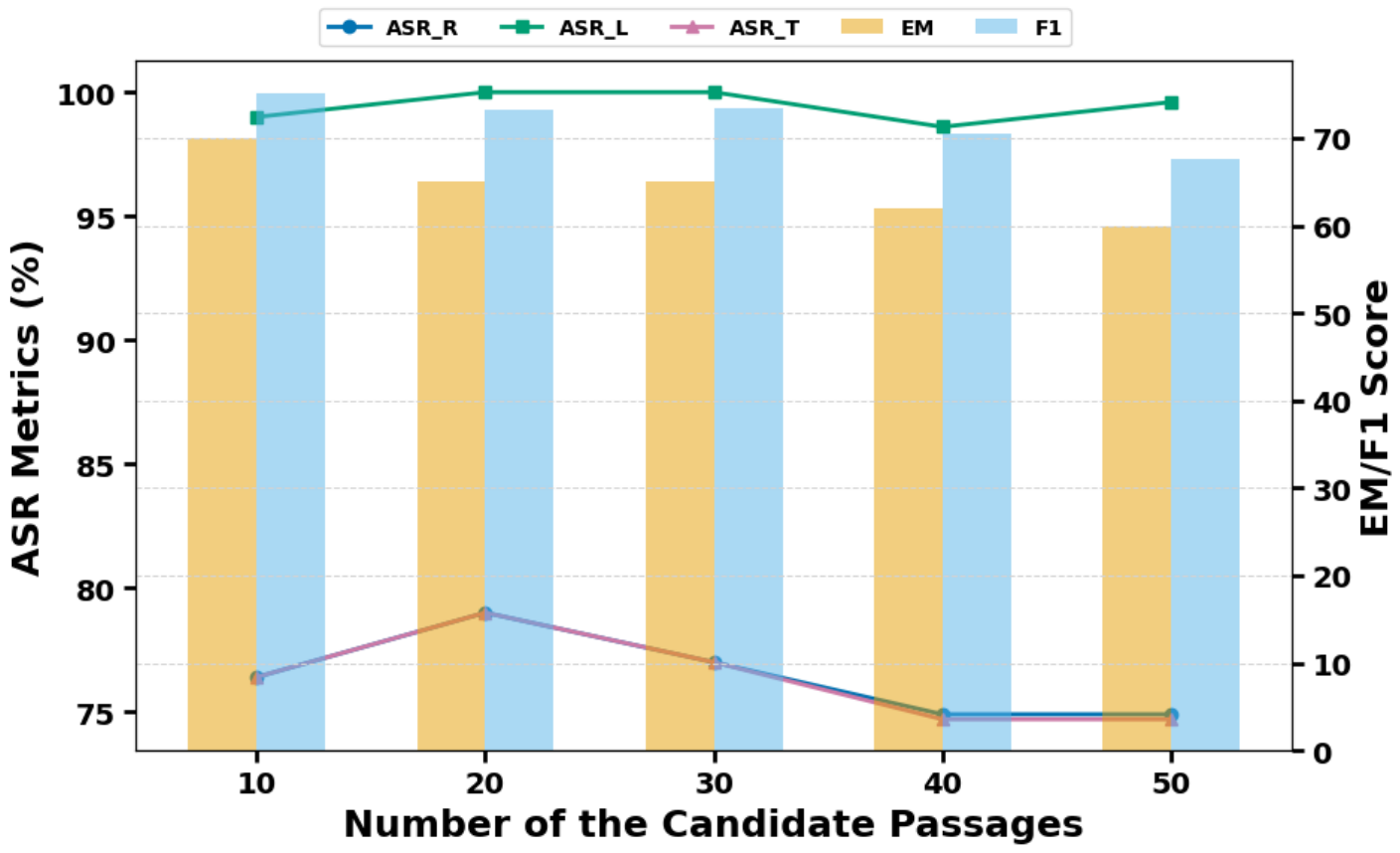}
    \caption{The performance of TPARAG under different sizes of malicious passage candidate sets.}
    \label{fig:analysis_candidate}
    \vspace{-2mm}
\end{figure}
Meanwhile, substitution rate also has a modest but consistent impact on end-to-end attack performance. Higher substitution rates are associated with lower EM scores, suggesting that increased token-level variability leads to more effective disruption of the RAG system’s answer generation. \\
\textbf{Optimized Iteration.} We further examine how the number of optimization iterations affects TPARAG's effectiveness.
Figure \ref{fig:analysis_iter} presents the performance trends across the iterations from $1^{st}$ to $5^{th}$ under the black-box RAG setting, following the setup in Section \ref{sec:04}.
The results show that the $ASR_R$ increases initially and peaks at the $3^{rd}$ iteration. Similarly, the $ASR_L$ exhibits an initial upward trend and stabilizes after the $3^{rd}$ iteration, indicating diminishing returns from further iterations. 
For the end-to-end attack performance, we observe a steady decline in answer accuracy with more iterations, indicating a consistent degradation of the RAG system’s ability to generate correct answers as malicious passages become more refined. \\
\textbf{The Number of Malicious Candidate Passages}.   
In TPARAG’s optimization attack stage, the size of the candidate set $\hat{D}_i$ may also impact the attack effectiveness.
Here, we experiment with candidate sets of size 10, 20, 30, 40, and 50 under the black-box settings, while keeping all other settings fixed.
As the experimental results show in Figure \ref{fig:analysis_candidate}, TPARAG achieves better end-to-end attack performance (i.e., lower EM and F1-Score) when using larger candidate sets. 
However, the results also reveal a decline in retrieval attack effectiveness as the candidate set size increases, with $ASR_R$ gradually decreasing beyond a size of 30. 
This trade-off is likely due to TPARAG’s optimization strategy, which emphasizes misleading the reader (i.e., minimizing the likelihood of generating the correct answer) while treating textual similarity (and thus retrievability) as a secondary evaluation factor.

\section{Conclusion}
We propose TPARAG, a novel attack framework that leverages a lightweight white-box LLM to perform token-level precise attacks on RAG systems. 
The framework is designed to be effective in both white and fully black-box RAG settings.
Moreover, we conduct extensive experiments using various LLMs as attackers, validating the effectiveness of our proposed attack method.

\section{Limitation}
While TPARAG demonstrates strong attack performance on both black-box and white-box RAG systems, there remains room for further improvement. 
First, the current experiments are limited by the choice of retriever, as only Contriever is used for both black-box and white-box RAG settings. 
Future work could extend the evaluation to a broader range of retrievers with diverse training objectives and architectures.
Second, due to computational constraints, our main experiments are conducted on datasets with hundreds of instances. Scaling the evaluation to larger datasets (e.g., thousands of queries) would further validate the robustness and generalizability of TPARAG.

\bibliography{custom}

\newpage
\appendix
\section{Appendix}
\label{appendix}
\subsection{TPARAG Algorithm}
Algorithm \ref{alg:TPARAG} illustrates the processing details of TPARAG's two-stage attack, where the malicious passages are iteratively optimized through the generation and optimization attack stages, starting from the initialization.
\begin{algorithm*}
\caption{TPARAG}
\label{alg:TPARAG}
\begin{algorithmic}
\Require Query $q$, Answer $a$, $m$ relevant document $D=\{d_1, d_2, ..., d_m\}$, Iterations $T$, Maximum Token Substitution Rate $pr_{sub}$, LLM attacker $LLM_{attack}$, Attack Locator $Loc$, Similarity metric $Sim$,
\State Compute the initial generation logits: $L=\{l_1, l_2, ..., l_m\}$ for $a$, $l_i:=P_{LLM_{attack}}(a|q, d_i; \theta_{LLM_{attack}})$,
\State Initialization threshold for generation attack: $l = \max_{l_i \in L} l_i$,
\State Compute the initial similarity score: $S:=\{s_1, s_2, ..., s_m\}$ for $q$, $s_i:=Sim(q, d_i)$,
\State Initialization threshold for retrieval attack: $s=\min_{s_i \in S} s_i$,
\State Entity-Based Attack Localization: $pos_{attack}:=Loc(a)$
\Loop{ $T$ times}
    \For{$i \in [0 \ldots m]$}
        \State $d'_i := LLM_{attack}(q \oplus d_i; \theta_{LLM_{attack}}) \in D'$
        \Comment{Generate parent malicious passage set $D'$ via $LLM_{attack}$}
        \For{$j \in [0 \ldots length(d'_i)]$}
            \State $T_{ij} := \mbox{Top-}k(P_{LLM_{attack}}(t_{1:j-1}, q, d_i))$
            \Comment{Compute/Save top-$k$  token substitution candidates for each generated token $t_{ij}$}
        \EndFor
    \EndFor
    \For {$i \in [0 \ldots m]$}
        \For {$j \in [0 \ldots length(d'_i)]$}
            \State Generate $pr_{rand}$
            \If{$t_{ij} \in pos_{attack}$ and $pr_{rand} < pr_{sub}$}
                \State $\hat{t}_{ij} := Random(T_{ij})$
                \Comment{Random select the replacement token}
            \Else
                \State $\hat{t}_{ij} := t_{ij}$
            \EndIf
        \State $\hat{d}_i := \hat{t}_{1:length(d'_i)}$
        \Comment{Reconstructed malicious passage candidate $\hat{d}_i$}
        \EndFor
    \State $\hat{s}_i:=Sim(\hat{d}_i, q)$, $\hat{l}_i:=P_{LLM_{attack}}(a|q, \hat{d}_i; \theta_{LLM_{attack}})$
    \EndFor
\State $\tilde{D}=\{\tilde{d}_i \mid i \in \text{Top-}m_{\text{ascending}}(\hat{l}_1, \dots, \hat{l}_m), Sim(q, \tilde{d}_i) > s, \hat{l}_i < l \}$
\Comment{Sort and find $m$ optimal malicious passages}
\EndLoop
\State \textbf{return:} Optimal malicious passage set $\tilde{D}$ for query $q$
\end{algorithmic}
\end{algorithm*}

\end{document}